# Analysing Long Short Term Memory Models for Cricket Match Outcome Prediction

**Rahul Chakwate**
Indian Institute of Technology Madras, India.
**Madhan R A**
American Express, Bangalore, India.

*Abstract* — As the technology advances, an ample amount of data is collected in sports with the help of advanced sensors. Sports Analytics is the study of this data to provide a constructive advantage to the team and its players. The game of international cricket is popular all across the globe. Recently, various machine learning techniques have been used to analyse the cricket match data and predict the match outcome as win or lose. Generally these models make use of the overall match level statistics such as teams, venue, average run rate, win margin, etc to predict the match results before the beginning of the match. However, very few works provide insights based on the ball-by-ball level statistics. Here we propose a novel Recurrent Neural Network model which can predict the win probability of a match at regular intervals given the ball-by-ball statistics. The Long Short Term Memory (LSTM) Model takes as input the ball wise features as well as the match level details available from the training dataset. It gives a prediction of winning the match at any time stamp during the match. This level of insight will help the team to predict the probability of them winning the match after every ball and help them determine the critical in-game changes they should make in their game strategies.[1]

*Keywords:* Cricket, ODIs, Machine Learning, Deep Learning, RNN, LSTM, Match Outcome Prediction.

## I. INTRODUCTION

Cricket is one of the most celebrated sports in the world. The One Day Internationals (ODIs) are the most watched cricket matches amongst all. They are held at a global level between the countries where each player plays for the country and represents the country. ODIs consist of two innings where each team bats for 50 overs each.

Due to the advancements in the instrumentation and sensor technologies, it has become easier to collect a lot of data from the match during the ongoing match. The love for the game and huge data availability has diverted the attention of the data analysts towards this game. Using the match data, sports analysts can provide useful insights on the match and recommendations to the team manager and its players. Machine learning consists of statistical methodologies where the model learns the representation of the data from the data itself. It is used to predict the dependent (unknown) variable in the dataset. The data scientists use these machine learning techniques to provide insights on the cricket match.

One of the most important intuitions researchers are curious about is the probability of a team winning the match. A lot of research is being conducted to accurately predict the winning team. Researchers achieve state-of-the-art results using various machine learning techniques including K-Nearest Neighbors, Logistic Regression, Support Vector Machines, Decision Trees, Random Forests, XGBoost and many more. Most of these works focus on predicting the match results at the beginning of the match using the available data about the teams, venue, pitch, weather, etc. collected over the years.

However, very few works are capable of modelling the ongoing match. Our LSTM based model can provide real time predictions of the match after every ball. It is capable of capturing the details of every ball and can give the win probability of a team based on the previous balls and overs. The win probability refers to the probability of the batting team winning the game at that instance. This level of prediction is necessary as there can be unforeseen circumstances happening during the match which can have a significant impact on the match results. Sometimes, a poorly performing team or a player can play excellently. This can be tracked by his previously played balls during the match. In this way, the results can change during the match. Our model can efficiently capture this information while other machine learning models may not. This type of model can be used in the recommendation system to the manager or the captain of the team about the critical in-game changes required to win the match.

## II. LITERATURE SURVEY

*A. Machine Learning Classification Techniques*

Several machine learning algorithms have been developed till date which have contributed to the evolution of data analytics and its vast applicability. These models learn from the intrinsic features of the data and can predict the suitable outcome. They can be broadly classified as regression models and classification models. However, to predict a binary outcome such as a win or lose, classification models are widely used. Some of the classification models are as follows:

1) KNN [1]: The simplest form of classifier is the K Nearest Neighbors classifier. These models transform the data into feature space and computes the k neighbors nearest to the test data point. It predicts the outcome of the test point according to the majority of the outcome of its neighbors.

2) Naive Bayes Classifier [2]: This statistical model is derived from the Bayes Theorem. It predicts the outcome based on the prior of the input variables. It assumes independence of feature and target variables.

3) Logistic Regression [3]: It is used for binary classification such as win/lose or head/tail. It applies a logistic function (such as sigmoid or tanh) to model the data.

4) Support Vector Machine [4]: They are the most widely used models for classification. They provide an optimal decision boundary which will separate the classes by the maximum margin and hence give a robust classification result.

5) Decision Tree Classifier [5]: It works on the principles of

---

[1] Code available at:
https://github.com/ruc98/American_Express_Ignite_2019_Challenge

a binary tree. The nodes act as a decision boundary separating points which are either side of a certain threshold. It breaks down the complex decision making task into multiple simple decision making tasks. It has the highest interpretability among all other complex classifiers.

6) Random Forest Classifier [6]: It is an ensemble created by cascading several decision tree classifiers in order to break down the complex classification task into simple tasks. It has proved to increase the accuracy as compared to the decision tree classifier but at the cost of its interpretability.

*B. Machine Learning in Cricket Analytics*

As one of the most popular sports in the world, cricket has gathered a lot of attention from the researchers and analysts. Predicting the winning team of the match is one of the most widely analysed topics. Machine learning being at the top of analytics, there is a lot of literature on the use of machine learning techniques to predict the outcome of the cricket match. Most of them use one or several of the above mentioned machine learning techniques for this purpose.

Shah et al. [7] used logistic regression on One Day Internationals (ODIs) Cricket dataset from the ICC to predict the winning team. They were able to achieve an accuracy of 81%.

Daniel et al. [8] used Decision Tree Classifier and XGBoost, a variant of it, to predict the Indian Premier League (IPL) matches. They got 94.8% accuracy in predicting the winner of the IPL match.

Jhanwar et al. [9] trained their models on the ODIs matches. They trained the classification models like KNN and logistic regression and could predict 71% of the matches accurately.

Gagana S. and K. Paramesha [10] made a comparative study of Naive Bayes, Decision Tree and Random Forest classifiers on the IPL matches. They trained the models with different train and validation splits and achieved around 90% accuracy on the dataset.

Rabindra L. et al. [11] analysed the past and the recent performances of the teams. Using Random Forest and Multiple Logistic Regression and using the past data, they could achieve satisfactory results.

Mayank et al. [12] devised the notion of Most Valuable Player (MVP) and found that it made a significant impact on the match results. He used Genetic Algorithms to create a recommendation system for the team captain in choosing his players and their positions.

Shimona et al. [13] used techniques from Data Mining to create a proper balance in the imbalanced dataset. They could achieve 97% accuracy on the balanced dataset. The performance reduced on the imbalanced data.

All the above papers tried to predict the results of the match before the beginning of the match using the past information. However, very few works focus on modelling the overs or balls of the match. One such work is by Jhawar et al. [14] which is the closest to our work. They attempt to model the ODI dataset. However, they train separate models for each over of the match which is very inefficient and unnecessary work. They first compare various machine learning techniques like Decision Tree, SVM, KNN, Gradient Boosting and Random Forest and find out that

Random Forest gives the best results. Then they use Random Forest to model the match data after every over. They achieve 75.68% of average accuracy across the overs. Our model, on the other hand, uses a single model to represent the entire match and yet can give match predictions after every ball of the match.

*C. Neural Networks and Recurrent Neural Networks*

Neural Networks[15] are a special type of machine learning algorithms. They attempt to model a complex function by mimicking the neurons in the human brain. Technically, they are simply multiple logistic regression cascaded on one another. However, neural networks have proved themselves useful in many complex data modelling tasks. Deep neural networks used in the natural language processing and computer vision domains have achieved state-of-the-art results on various tasks. Some of the models even surpass the human capabilities in completing a vision based or language based task. Recurrent Neural Networks (RNNs) [16] are a special type of neural networks where connections between the nodes are along a temporal sequence. Long Short Term Memory (LSTM) [17] models are the most widely used RNNs because of their selective gating and direct gradient transfer properties. They are used to accomplish any kind of temporal or sequential task such as machine translation, weather forecast, speech recognition, Protein Sequence modelling and many more. In the game of cricket, the ball-by-ball data can also be considered as a time sequence. Hence, we find it useful to model this data with the help of LSTMs.

### III. OUR METHODOLOGY

The workflow diagram of our methodology is given in the Fig. 1 below.

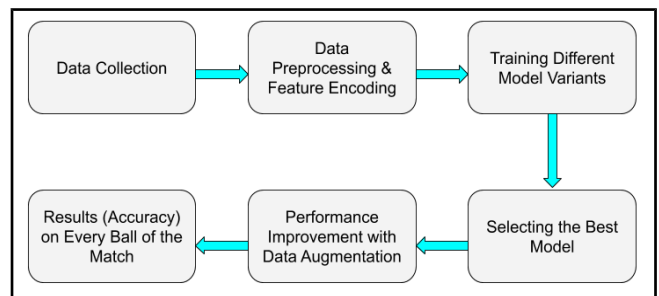

Fig. 1. Workflow graph.

*A. Dataset*

The dataset used in our experiments is the open source ODIs (One Day Internationals) dataset. It consists of all the major international cricket matches from 1971 onwards. It contains 1597 matches with 50 overs per innings. Each of the matches has ball-by-ball level statistics of both the innings of the match. The ball attributes and their description are mentioned in Table I.

Table I. Feature description

| Attributes of ball | Description |
|---|---|
| Innings number | First inning or Second inning |
| Over and Ball | Number of the over and the ball |
| Batting Team Name | Team batting during that ball |

| Batsman Name | Batsman batting at that ball |
|---|---|
| Non-striker Name | Non-strike batsman on the field |
| Bowler Name | Bowler of that ball |
| Runs-off-bat | Runs scored in that ball |
| Extras | Extra runs during no or wide ball |
| Wicket | 1 if there is any wicket in that ball |

The dataset also has the following match level statistics: team, gender, season, date, series, match number, venue, city, toss winner, toss decision, player of the match, umpire, tv umpire, match referee, winning team and runs scored by the winning team.

*B. Dataset Preprocessing and Encoding*

For our LSTM model, we use the ball by ball level data while for the machine learning model, we use the match level statistics as explained later in this section. However, the data needs to be preprocessed before feeding it into the neural network. The over and balls are given in the over.ball format where ball is given from 1 to 6 (or 7 if no ball). Ideally, there must be 300 balls in one inning of the ODI. We add any extra runs from no ball to its previous ball to form a vector of constant length of 300 balls. Next, for each of the balls, the continuous data is used as it is in the dataset. For the categorical data such as names, we form a list of all the team and player names and remove the outliers in the data such as a team with a single match. Then we form a One-Hot encoding of the categorical data. For instance, the list of 30 teams is converted into 30 vectors where the entry is 1 if the team is present and zero otherwise.

*C. Our Framework Overview*

The architecture of our framework is given in Fig. 2. It shows the best variant of our model. The model consists of three blocks: the Input Transformation (IT) block, the LSTM block which is the main building block and the Output Transformation (OT) block.

The IT block takes a single ball's feature vector as input and transforms it into a higher dimensional feature vector. This block is shared by all the balls in that innings. The output of this layer is fed to the LSTM cell where the current cell's weights are influenced by the corresponding input features of the same timestamp and the output features of the previous timestamp. Next, the output of the LSTM cell is fed to the OT block which transforms the LSTM cell output from a higher dimension to a two dimensional vector (win or lose node). The softmax of this output layer gives the win probabilities of the match.

Fig. 2a. Our all inputs single output model.
Fig. 2b. Our All inputs all outputs model.

The final two node output is compared with the ground truth of win or lose by the Binary Cross Entropy Loss. This loss is aggregated for all the balls of the innings and the mean loss is used to train the network.

*D. Network Variants: Four different types of models*

We propose four variants of our network and conduct rigorous experimentation to quantitatively compare between the networks. Our variants are:

1) All overs input and single output model (model A): The architecture of this model is depicted in Fig. 2a. As the name suggests, the LSTM cell has an input channel to all the timestamps but the output channel is present only at the final cell. In this model, only one prediction is possible at the end of the match. One needs to train separate models for predicting the match after every ball.

Say we need to predict the match after the j$^{th}$ ball. Then all the data fed to the network need to be up to the j$^{th}$ ball. The output node will be attached to the j$^{th}$ timestamp and that gives the prediction of the model after the j$^{th}$ ball.

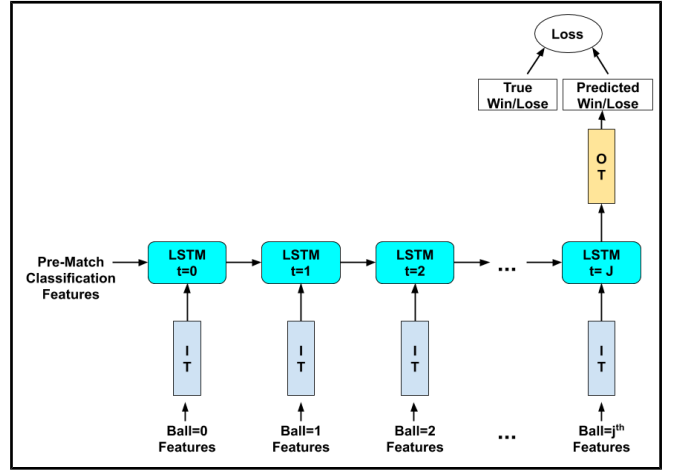

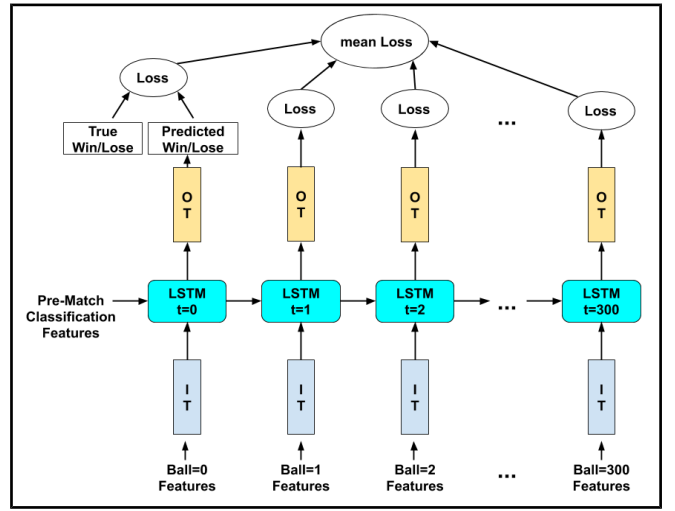

2) All overs input and multi output model (model B): The architecture of this model is shown in Fig. 2b. It contains all the inputs and all the outputs to the LSTM cell. Hence, one can predict the match at every ball of the game by looking at the output node of the corresponding timestamp.

3) Randomly sampled input and single output model (Model C):

This model is similar in structure as that of Model A. Except that the input to this model would be a randomly sampled sequence among all sequences available. To illustrate this model, consider the example below.

Let's say we have 1000 matches in the training data. We split the second innings ball by ball – leading us to a maximum of 1000*300.. Then we set each of these records to predict the match was won or not.
M1B1 – W
M1B1,M1B2 – W
M1B1,M1B2,M1B3... - W
M2B1 – L
M2B1,M2B2 – L
Until M1000

Where M is the match and B is the ball and each combination of MB is a unique record. By this way you can train the RNN to predict at any stage of the chase well.

4) Cumulative sum of ball features as input and single output model (Model D):

This model is similar in structure as Model A. Instead of feeding data for each ball, we feed the cumulative data up to that ball. By doing this, the model would be able to efficiently compare the cumulative runs scored till that ball to that of the target. It will also learn to model the effects of the pressure built on the batsman when the number of wickets fallen are more.

The experiments conducted with the above four model variants are shown in the experiments section below.

### IV. EXPERIMENTS AND ABLATION STUDY

*A. Performance Comparison of the Model Variants:*

The four model variants are compared in this section. Table II. reports the test accuracy in % on the ODIs dataset. Models A,B,C and D are as explained in the section above. "**J**" is the number of balls bowled till then. So in short, the table gives the accuracy of the predictions of the model after the **J**$^{th}$ ball.

Table II. Model Comparison

| Model | Test Accuracy (%) | | |
|---|---|---|---|
| | J=300 | J=250 | J=200 |
| Model A | 93.7 | 63.5 | 57.2 |
| **Model B** | **95.7** | **75.1** | **66.4** |
| Model C | 79.6 | 66.7 | 59.2 |
| Model D | 60.8 | 52.6 | 52.6 |

As seen in Table II, the multi outputs model (model B) performs the best on the test data. It converges to the solution in 100 epochs which is faster than the other models. Henceforth, we decide to use this architecture for further experiments.

*B. Pre-match Classification*

We experiment on the match predictions before the match as done by most other papers using machine learning techniques which use match level statistics like the venue, weather, pitch, etc. We experiment with AdaBoost and Light Gradient Boosting Mechanism (LGBM) methods which are a variant of Decision Trees with boosting. The results are shown in Table III. LGBM clearly performs better than the other method.

Table III: Pre-match Classification Results

| Model | Train Accuracy | Test Accuracy |
|---|---|---|
| AdaBoost | 61.0% | 47.2% |
| LGBM | 76.3% | 61.3% |

*C. Performance improvement*

We conduct an ablation study of our model. In Table IV, first we show our models improvement by augmenting the pre-match results from the subsection *B* of this paper. We append the results of LGBM to the first ball's input features and pass it to the LSTM network. In this way we get an average performance boost of about 2%. Further we append the target score from the first innings of the match to every input of the network. This itself gives around a 13% average performance boost to our network. This suggests that target score from the previous inning is a very important factor in predicting the match outcome at every ball. The model internally learns how to effectively use this target score to improve the predictions. We also append the number of wickets fallen in the first innings. It seems that this data is also useful for the model as it further gives around a 2% average boost to the model.

Table IV: Improved Performance Results

| Model | Test Accuracy (%) | | |
|---|---|---|---|
| | J=300 | J=250 | J=200 |
| Baseline | 95.7 | 75.1 | 66.4 |
| Augmenting Pre-match results | 96.3 | 77.6 | 68.2 |
| Augmenting Target Scores | 98.0 | 90.1 | 87.1 |
| **Augmenting Wickets** | **98.4** | **92.5** | **88.2** |

*D. Best Model and its performance with overs*

We experiment on four different model variants and select the best one in the subsection *A*. We also experiment by augmenting additional information to the model from the first innings and the pre-match classification model in subsections *B* and *C*. Finally, we get our best performing model. We report the performance of this model after every ball of the match with the graph shown in Fig. 3. It shows the test accuracy plotted against the number of balls in the match. We also report the train and test accuracies at some of the intermediate overs in Table V.

Table V: Best Model Results

| Balls | Train Accuracy (%) | Test Accuracy (%) |
|---|---|---|
| 300 (50 overs) | 98.3 | 98.4 |
| 240 (40 overs) | 91.0 | 92.1 |
| 180 (30 overs) | 88.7 | 88.0 |
| 120 (20 overs) | 85.2 | 87.3 |
| 90 (15 overs) | 80.4 | 85.6 |
| 60 (10 overs) | 78.1 | 82.7 |
| 30 (5 overs) | 70.2 | 72.4 |
| 6 (1 over) | 62.3 | 62.7 |

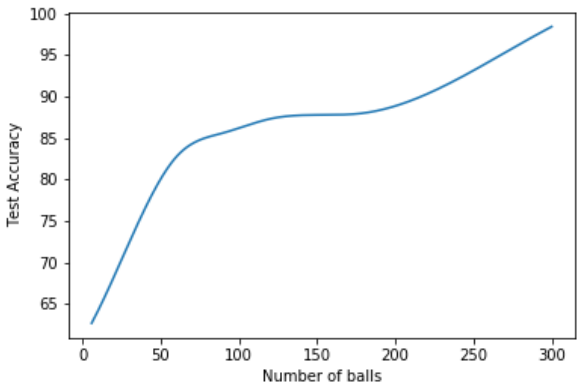

Fig. 3. Test Accuracy variation of our best model after the number of balls played in the match.

V. CONCLUSION AND APPLICATIONS

The key achievements of our model are as follows:
- Right from the 10th Over, it predicts 82.7% of the matches correctly into win or lose!
- The accuracy of predicting the match increases further to 87.3% at the 20th Over and 88.0% at the end of 30th Over. Which increases to 98.4% at the end of the match.
- In this way, our model can predict the outcome of the match at any instance of the match. The accuracy increases as the game reaches its end.
- This observation is very straightforward and intuitive. As we reach the end of the game it is very easy to predict the winning team by simply looking at the target score and the current runs. The model automatically learns this relationship between the runs and the target score.

We also observe that, at the end of the match, the model should be predicting 100% results as correct given the target score of first innings and all the runs in the second innings but the model fails to capture this trivial result. One reason for this is the abandoned matches. This can be due to unforeseen situations like the rains which resulted in deciding the winning team by some other algorithm like Duckworth Lewis Method rather than the general rules of the game. Other reasons could be the match ties in which case the win is ambiguous. However, except these matches, the model predicts almost all other matches correctly in the end.

Our model can be deployed in practice to predict matches after the 10th Over as the accuracies are impressive after the first 10 overs. It goes from 82.7% in the 10th over to 98.4% at the end of the game. These numbers are very promising and can be helpful in designing a recommendation system to the team manager or the team captain in making critical decisions during the game.